\documentclass[10pt,twocolumn,letterpaper]{article}

\usepackage{cvpr}
\usepackage{times}
\usepackage{epsfig}
\usepackage{graphicx}
\usepackage{amsmath}
\usepackage{amssymb}
\usepackage{subfigure}
\usepackage{array}
\usepackage{multirow}
\usepackage{diagbox}
\usepackage{graphicx}
\usepackage{booktabs}
\usepackage{gensymb}
\usepackage{xspace}
\usepackage[breaklinks=true,bookmarks=false]{hyperref}
\usepackage{cleveref}
\newcommand{\unit}[1]{\ensuremath{\,\mathrm{#1}}}
\newcommand{\OURS}{SSCNet\xspace}
\newcommand{\mypara}{\vspace*{-3mm}\paragraph}
\DeclareMathOperator{\sign}{sign}

\cvprfinalcopy % *** Uncomment this line for the final submission

\def\cvprPaperID{646} % *** Enter the CVPR Paper ID here

\begin{document}

%%%%%%%%% TITLE

%\title{SceneFill: Simultaneous Volumetric Completion and Semantic Labeling}
%\title{Volumetric Semantic Scene Completion Network} 
%\title{\OURS for Simultaneous 3D Scene Completion and Labeling} 
\title{\OURS for  Simultaneous Scene Completion and Semantic Labeling}
\title{Semantic Scene Completion from a Single Depth Image}
%\title{Simultaneous Volumetric Scene Completion and Labeling} 
%\title{SceneVox: Simultaneous Volumetric Scene Completion and Labeling}
%\title{SceneVox: A Deep Volumetric Representation for Simultaneous Scene Completion and Labeling}

%for Volumetric Amodal Scene
%\title{SceneVox: Volumetric Amodal Scene Segmentation from Single View Depth} %for Volumetric Amodal Scene  
%\title{Semantic Scene Completion: labeled voxels from single view depth images} 
\author{Shuran Song \quad Fisher Yu \quad  Andy Zeng \quad  Angel X. Chang \quad  Manolis Savva \quad  Thomas Funkhouser \\
Princeton University}

\maketitle
%\thispagestyle{empty}

%%%%%% ABSTRACT %%%%%%
\begin{abstract}
This paper focuses on semantic scene completion, a task for producing a complete 3D voxel representation of volumetric occupancy and semantic labels for a scene from a single-view depth map observation.  Previous work has considered scene completion and semantic labeling of depth maps separately.  However, we observe that these two problems are tightly intertwined.
To leverage the coupled nature of these two tasks, we introduce the semantic scene completion network ({\bf \OURS}), an end-to-end 3D convolutional network that takes a single depth image as input and simultaneously outputs occupancy and semantic labels for all voxels in the camera view frustum.
Our network uses a dilation-based 3D context module to efficiently expand the receptive field and enable 3D context learning.
To train our network, we construct SUNCG - a manually created large-scale dataset of synthetic 3D scenes with dense volumetric annotations.
Our experiments demonstrate that the joint model outperforms methods addressing each task in isolation 
and outperforms alternative approaches on the semantic scene completion task. 
The dataset, code and pretrained model will be available online upon acceptance.
\end{abstract}
%We focus on the task of semantic scene completion, which aims to produce both volumetric occupancy and semantic labels for the full 3D space given a single view depth map observation of a 3D scene. We observe that the occupancy patterns of the environment and the semantic labels of the objects are tightly intertwined.
\vspace{-2mm}
%%%%%%%%%%%%%%%%%%%%%%%%%%%%%%%%%%%%%%%%%% Introduction %%%%%%%%%%%%%%%%%%%%%%%%%%%%%%%%%%%%%%%%%%
\section{Introduction}

%why important
We live in a 3D world where empty and occupied space is determined by the physical presence of objects. To successfully navigate within and interact with the world, we rely on an understanding of both the 3D geometry and the semantics of the environment. Similarly, for a robot, the ability to infer complete 3D shape from partial observations is necessary for low-level tasks such as grasping and obstacle avoidance \cite{Grasping}, while the ability to infer the semantic meaning of objects in the scene enables high-level tasks such as retrieval of objects.

\begin{figure}[t]
\vspace{-5mm}
    \includegraphics[width=\linewidth]{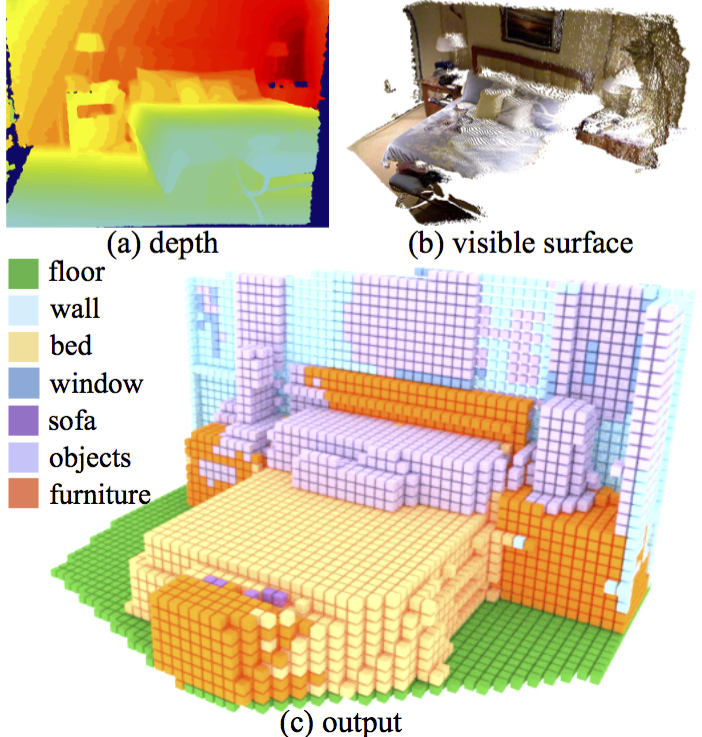}
    \caption{{\bf Semantic scene completion.} (a) Input single-view depth map (b) Visible surface from the depth map; color is for visualization only. (c) Semantic scene completion result: our model jointly predicts volumetric occupancy and object categories for each of the 3D voxels in the view frustum. Note that the entire volume occupied by the bed is predicted to have the bed category.}
    \label{fig:task}
    \vspace{-3mm}
\end{figure}

\begin{figure}[t]
\vspace{-5mm}
    \includegraphics[width=\linewidth]{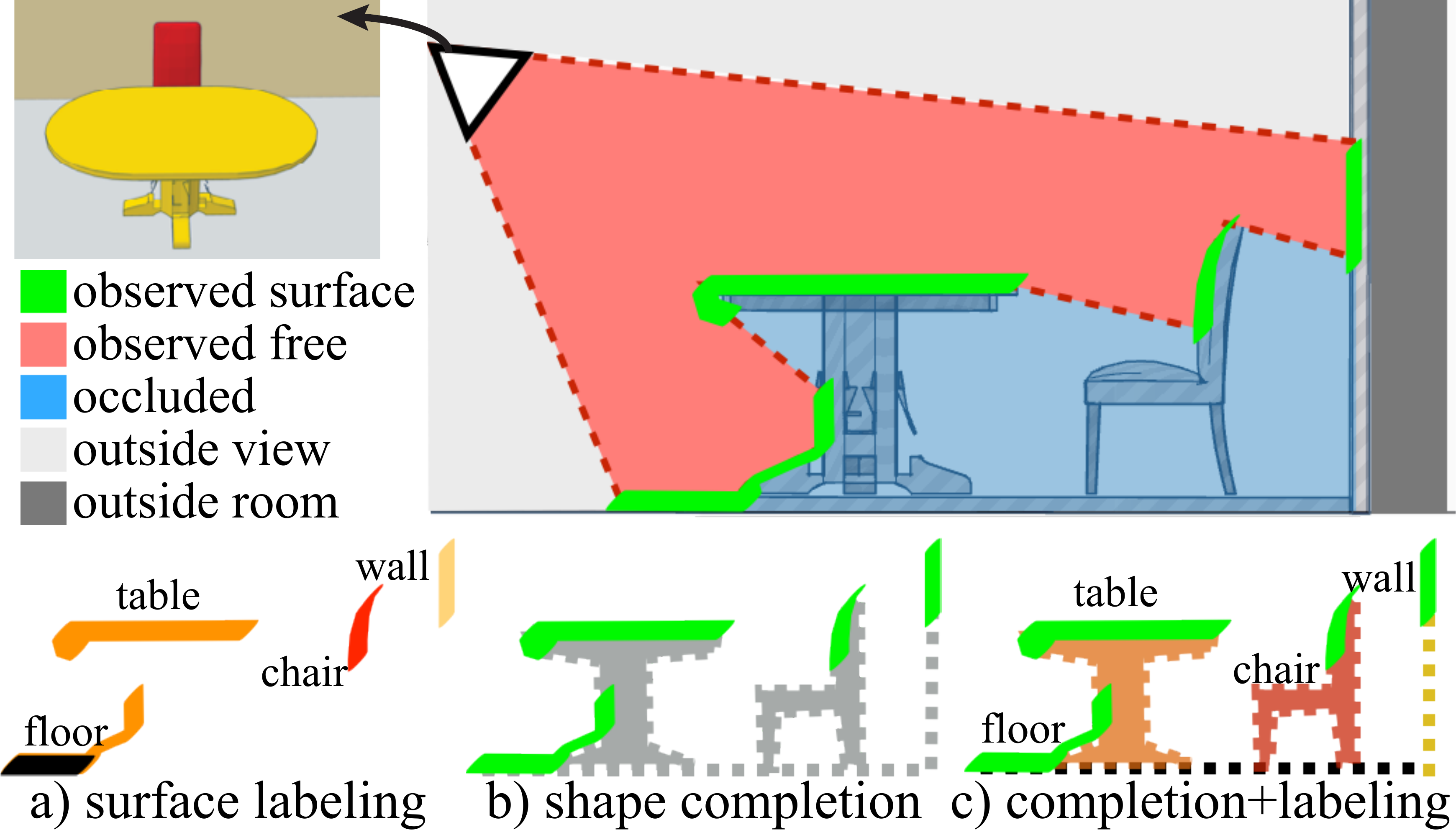}
    \caption{Given a single-view depth  observation of a 3D scene the goal of our \OURS is to predict both occupancy and object category for the voxels on the observed surface and occluded regions. }
    \label{fig:space}
    \vspace{-3mm}
\end{figure}

%zpp goal + key observation 
With this motivation, our goal is to have a model that predicts both volumetric occupancy (i.e., scene completion) and object category (i.e., scene labeling) from a single depth image of a 3D scene --- in this paper we refer to this task as {\bf semantic scene completion} (\Cref{fig:task}). 
Prior work is limited to address only part of this problem as shown in \Cref{fig:space}: RGB-D segmentation approaches consider only visible surfaces without the full 3D shape \cite{guptaCVPR13,ren2012rgb}, while shape completion approaches consider only geometry without semantics \cite{FirmanCVPR2016} or a single object out of context \cite{Nguyen_2016_CVPR,3DShapeNets}.

%zpp our insight 
Our key observation is that the occupancy patterns of the environment and the semantic labels of the objects are tightly intertwined. Therefore, the two problems of predicting voxel occupancy and identifying object semantics are strongly coupled.  In other words, knowing the identity of an object helps us predict what areas of the scene it is likely to occupy without direct observation (e.g., seeing the top of a chair behind a table and inferring the presence of a seat and legs).  Likewise, having an accurate occupancy pattern for an object helps us recognize its semantic class.

% approach overview
To leverage the coupled nature of the two tasks we jointly train a deep neural network using supervision targeted at both tasks. Given a single-view depth map as input, our semantic scene completion network ({\bf \OURS}) produces one of N+1 labels for all voxels in the view frustum.
Each voxel is labeled as occupied by one of N object categories or free space.
Most critically, this prediction extends beyond the projected surface implied by the depth map, thus providing occupancy information for the entire scene.

%zpp why challenging / not yet solved: big context, data
To achieve this goal there are several issues that must be addressed. First, how do we effectively capture contextual information from 3D volumetric data, where the signal is sparse and lacks high frequency detail?
Second, since existing RGB-D datasets only provide annotations on visible surfaces, how do we obtain training data with complete volumetric annotations at scene level?

%To capture the {\bf 3D context},
To address the first issue, we design a 3D dilation-based context module that efficiently expands our network's receptive field to model the contextual information.
We find that a big receptive field is crucial for the task. As demonstrated in \Cref{fig:space}, looking at the small region of a chair in isolation, it is hard to recognize and complete the chair. However, if we consider the context due to surrounding objects, such as the table and floor, the problem is much easier.

To address the  second issue, we construct SUNCG, a large-scale synthetic 3D scene dataset with more than 45622 indoor environments designed by people.
All the 3D scenes are composed of individually labeled 3D object meshes, from which we can compute 3D scene volumes with dense object labels though voxelization.

Our experiments with these solutions demonstrate that a method that jointly predicts volumetric occupancy and object semantic can outperform methods addressing each task in isolation. Both the 3D context model learned by our network and the large-scale synthetic training data help to improve performance significantly.

%zpp results and contributions 
Our main contribution is to formulate an end-to-end 3D ConvNet model (\OURS) for the joint task of volumetric scene completion and semantic labeling.
In support of that goal, we design a dilation-based 3D context module that enables efficient context learning with large receptive fields.
To provide the training data for our network, we introduce SUNCG, a manually created large-scale dataset of synthetic 3D scenes with dense occupancy and semantic annotations.

\begin{figure*}[t]
 \vspace{-6mm}
    \includegraphics[width=\linewidth]{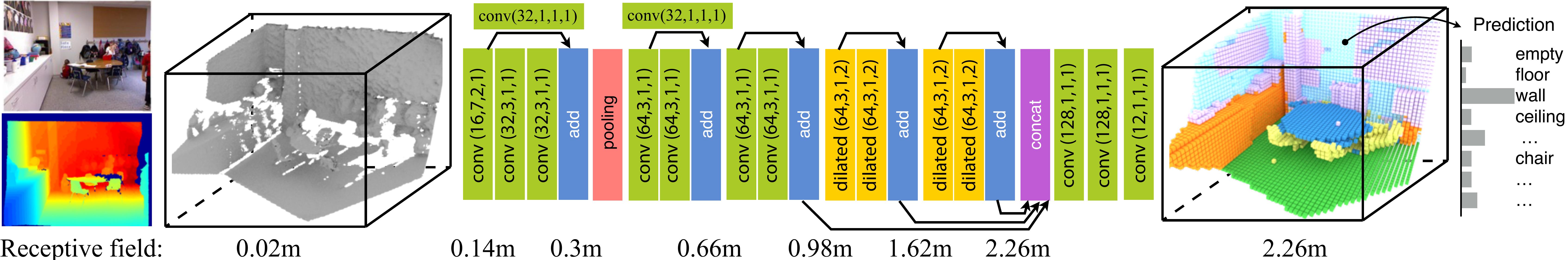}
     \caption{{\bf \OURS: Semantic scene completion network.} Taking a single depth map as input, the network predicts occupancy and object labels for each voxel in the view frustum. The convolution parameters are shown as (number of filters, kernel size, stride, dilation).}
     %, where dilation equal to one indicates normal 3D convolution
     \label{fig:network}
     \vspace{-3mm}
\end{figure*}

%%%%%%%%%%%%%%%%%%%%%%%%%%%%%%%%%%%%%%%%%% RELATED WORK %%%%%%%%%%%%%%%%%%%%%%%%%%%%%%%%%%%%%%%%%%
\section{Related work}
We review related work on RGB-D segmentation, 3D shape completion, and voxel space semantic labeling.

\mypara{RGB-D semantic segmentation.}
% that only reasons with the visible surfaces
%Many methods have been proposed for RGB-D segmentation
Many prior works focus on RGB-D image segmentation
\cite{guptaCVPR13,ren2012rgb,NYUdataset,lai2014unsupervised}.
However, those methods focus on obtaining semantic labels for only the observed pixels without considering the full shape of the object, and hence cannot directly perform scene completion or predict labels beyond the visible surface.
%treat the depth maps as extra channel(s) of a color image

\mypara{Shape completion.}
Other prior works focus on single object shape completion \cite{Grasping,rock2015completing,3DShapeNets,Nguyen_2016_CVPR}.
To apply those methods to scenes, additional segmentation or object masks would be required.
For scene completion, when the missing regions are relatively small, methods using plane fitting \cite{monszpart2015rapter} or object symmetry \cite{kim2012acquisition,mattausch2014object} can be applied to fill in holes. However, these methods heavily rely on the regularity of the geometry and often fail when the missing regions are big.
Firman \etal\cite{FirmanCVPR2016} show promising completing results on scenes.
However, their approach is based purely on geometry without semantics, and thus it produces less accurate results when the scene structure becomes complex. 

\mypara{3D model fitting.}
One possible approach to obtain the complete geometry and semantic labels for a scene is to retrieve and fit instance-level 3D mesh models to the observed depth map \cite{guptaCVPR15,SlidingShapes,Geiger2015GCPR,Lai10,Nan12,Li15,Kim12acquiring}.
However, the prediction quality of this type of approach is limited by the quality and variety of 3D models available for retrieval.
Naturally, observed objects that cannot be explained by the available models tend to be missed.   
Or, if the 3D model library is large enough to include all observations, then a difficult retrieval and alignment problem must be solved.
Alternatively, it is possible to use 3D primitives such as bounding boxes to approximate the 3D geometry of objects \cite{RGBDcuboid,lin2013holistic,DSS}.
However, the bounding box approximation limits the geometric detail of the output predictions.
%Additional global reasoning about object arrangements and room layouts can be applied to obtain more physically plausible results \cite{Geiger2015GCPR}.
%In addition, retrieval methods typically rely on first predicting semantic labels and occupancy, which is the task addressed by this paper.
%Geiger el al.~\cite{Geiger2015GCPR} use a total of 38 mesh models in 21 object categories.

\mypara{Voxel space reasoning.}
Another line of work completes and labels 3D scenes, but with separate modules for feature extraction and context modeling.
Zheng \etal~\cite{ZhengZYIZ13Physics} predict the unobserved voxels by physical reasoning. %under a Manhattan world assumption.
Kim \etal~\cite{kim20133d} train a Voxel-CRF model from labeled floor plans to optimize the semantic labeling and reconstruction for indoor scenes.
Hane \etal~\cite{HaneZCAP13} and Blaha \etal \cite{blahalarge} use joint optimization for multi-view reconstruction and segmentation for outdoor scenes.
However, this line of work uses predefined features, and separates the feature learning from the context modeling, and it is expensive for CRF-based models to encode long-range contextual information.
In contrast, our model is able to jointly learn the low-level feature representation and high-level contextual information end-to-end from large-scale 3D scene data, directly modeling long-range contextual cues though big receptive field.

\mypara{Learning from synthetic scene data.}
Our paper leverages data generated from a large-scale synthetic 3D scene dataset. 
Although recent works have been focusing on generating segmentation labels for 2D image through rendering synthetic scenes \cite{SceneNet,Richter_2016_ECCV}, the 3D aspect of such data has not been fully utilized.
Existing datasets focus either on objects \cite{shapenet2015,3DShapeNets} or a small number of rooms (57 rooms in \cite{SceneNet}). In contrast, our dataset is several orders of magnitude larger than existing 3D scene datasets (45,622 houses with 775,574 rooms) providing a diverse set of furniture arrangements manually created by people.

%%%%%%%%%%%%%%%%%%%%%%%%%%%%%%%%%%%%%%%%%% Approach %%%%%%%%%%%%%%%%%%%%%%%%%%%%%%%%%%%%%%%%%%
\section{Semantic scene completion network}

Given a single-view depth map observation of a 3D scene, the goal of our semantic scene completion network is to map the voxels in the view frustum to one of the class labels $C = \{c_{0}, ... c_{N+1}\}$, where N is number of object classes and $c_{0}$ represents empty voxels.
During training, we render depth maps from virtual viewpoints of our synthetic 3D scenes and voxelize the full 3D scenes with object labels as ground truth. During testing, the observation depth images come from a RGB-D camera.

\Cref{fig:network} shows an overview of our processing pipeline.
We take a single depth map as input and encode it as a 3D volume.
This 3D volume is then fed into a 3D convolutional network, which extracts and aggregates both local geometric and contextual information.
The network produces the probability distribution of voxel occupancy and object categories for all voxels inside the camera view frustum.

The following subsections describe the core issues addressed in the design of the system: the data encoding (\Cref{sec:encode}), network architecture (\Cref{sec:network}) and training data generation (\Cref{sec:dataset}).

\begin{figure*}[t]
 \vspace{-5mm}
(a) Object centric networks \cite{3DShapeNets,VoxNet}~~~~~~~~~~(b) 3DMatch \cite{3DMatch} ~~~~~~~~~~ (c) Deep Sliding Shape \cite{DSS} ~~~~~~~~ (d) \OURS ~~~~~~~~~~~~~~~~~~\\
\centering
\includegraphics[width=1\linewidth]{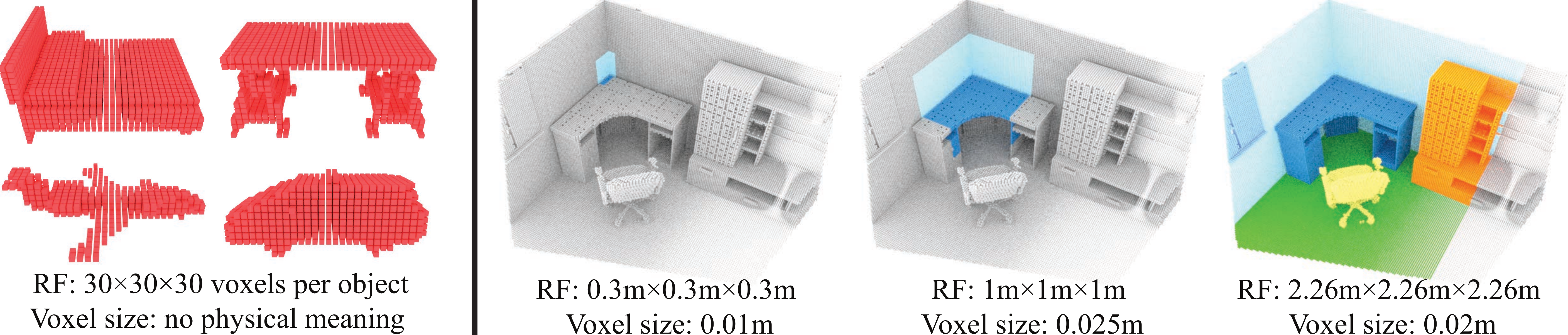}

 \caption{{\bf Comparison of receptive fields and voxel sizes between \OURS\ and prior work.}
 (a) Object centric networks such as \cite{3DShapeNets} and \cite{VoxNet} scale objects into the same 3D voxel grid thus discarding physical size information.
 In (b)-(d), colored regions indicate the effective receptive field of a single neuron in the last layer of each 3D ConvNet.  With the help of 3D dilated convolution \OURS drastically increases its receptive field compared to other 3D ConvNet architectures \cite{DSS,3DMatch} thus capturing richer 3D contextual information.}
 \label{fig:RF_compare}
  \vspace{-3mm}
\end{figure*}

\begin{figure}[t]
    \includegraphics[width=\linewidth]{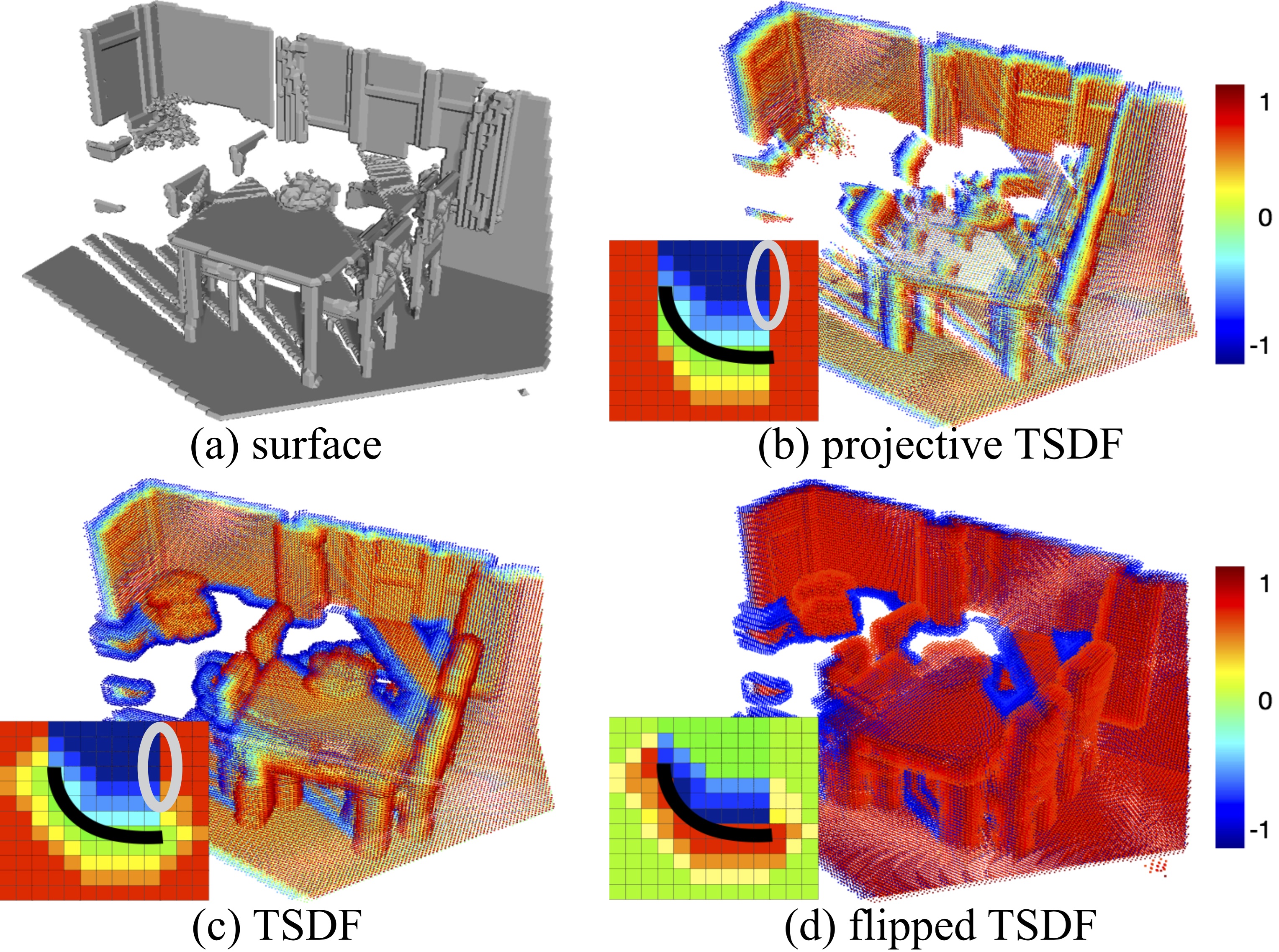}
    \caption{{\bf Different encodings for surface (a).} The projective TSDF (b) is computed with respect to the camera and is therefore view-dependent.
    The accurate TSDF (c) has less view dependency but exhibits strong gradients in empty space along the occlusion boundary (circled in gray). In contrast, the flipped TSDF (d) has the strongest gradient near the surface.}% providing more meaningful signal for the network.}% to learn geometry feature.}
    \label{fig:tsdf}
    \vspace{-3mm}
\end{figure}

\subsection{Volumetric data encoding}
\label{sec:encode}
The first issue we need to address is how to encode the observed depth as input to the network. 
For the semantic scene completion task, the ideal encoding should directly represent the 2D observation into the same 3D physical space as the output in a way that is invariant to the viewpoint projection, and provide a meaningful signal for the network to learn geometry and scene representation. 
To this end, we adopt Truncated Signed Distance Function (TSDF) to encode the 3D space, where every voxel stores the distance value $d$ to its closest surface, and the sign of the value indicates whether the voxel is in free space or in occluded space.
To better suit our task, we make the following modifications to the standard TSDF.
% that clearly differentiate observed surface, observed free space and occluded region.
%In our task, the observation is a 2D array of depth values with respect to the camera, while the output is a 3D voxel grid class labels from $C$.To bridge the gap between input and output dimensionality, we represent the data as a dense 3D volume that directly encodes the 2D observation into the same 3D physical space as the output.
%This 3D volumetric representation encodes the data in real world dimensions, and is therefore invariant to viewpoint projection. 

\mypara{Eliminate view dependency.}
Most RGB-D reconstruction pipelines speed up the TSDF computation by using the projective TSDF which finds the closest surface points only in the line of sight of the camera \cite{kinectfusion}.
This projective TSDF is fast to compute, but is inherently view-dependent.
%We want a representation with reduced view dependency, so 
Instead, we choose to compute the distance to the closest point anywhere on the full observed surface.

\mypara{Eliminate strong gradients in empty space.}
Another issue with TSDF is that strong gradients occur in the empty space along the occlusion boundary between $\pm d_\mathrm{max}$. %, which most of the time correspond to empty voxels.
It is possible to eliminate this gradient by removing the sign, however, the sign is important for completion task since it indicates the occluded regions of the scene that need to be completed.
To solve this problem we flip the TSDF value $d$ as follows:
$d_\mathrm{flipped} = \sign(d)(d_\mathrm{max}-d)$.
This flipped TSDF has the strong gradient on surface, providing a more meaningful signal for the network to learn better geometric features.
The different encoding is visualized in \Cref{fig:tsdf}, and
\Cref{table:control} shows its impact on performance.
%We show in our experiments that the type of TSDF encoding has an impact on our results.

\begin{figure*}[t]
 \vspace{-5mm}
    \includegraphics[width=\linewidth]{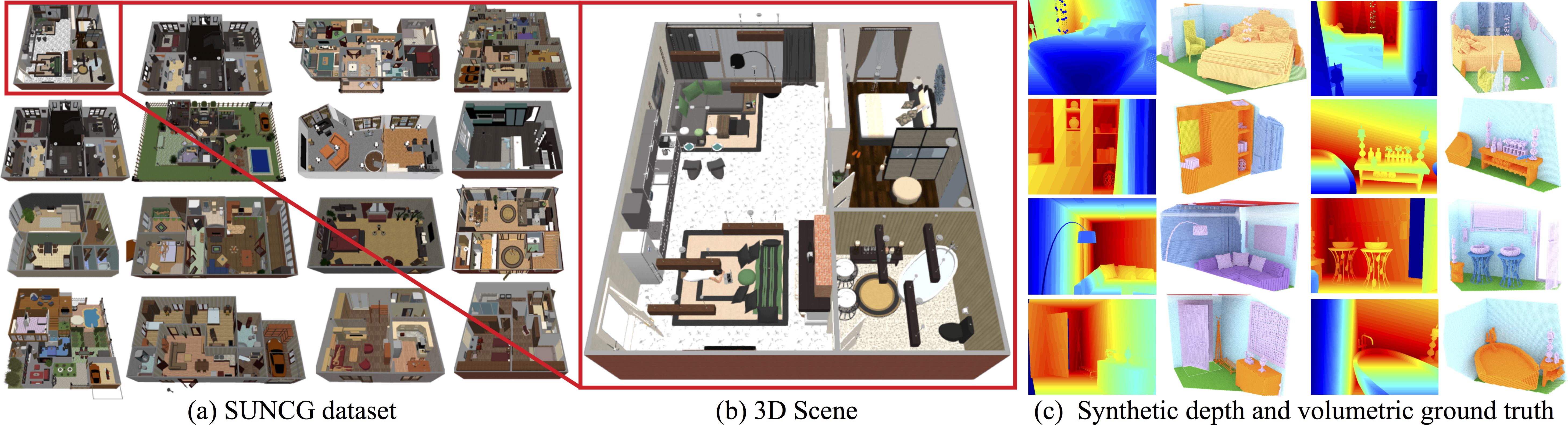}
    \caption{{\bf Synthesizing Training Data.}
    We collected a large-scale synthetic 3D scene dataset to train our network.
    For each of the 3D scenes, we select a set of camera positions and generate pairs of rendered depth images and volumetric ground truth as training examples.}%we render it from hundreds of view angles to generate a pool of positive training
    \label{fig:data}
    %\vspace{-3mm}
\end{figure*}

\subsection{Network architecture}
\label{sec:network}

The network architecture of \OURS is shown in \Cref{fig:network}.
Taking a high-resolution 3D volume as input, the network first uses several 3D convolution layers to learn a local geometry representation.
We use convolution layers with stride and pooling layers to reduce the resolution to one fourth of original input.
Then, we use a dilation-based 3D context module to capture higher-level inter-object contextual information.
After that, the network responses from different scales are concatenated and fed into two more convolution layers to aggregate information from multiple scales. 
At the end, a voxel-wise softmax layer is used to predict the final voxel label.
Several shortcut connections are added for better gradient propagation.  
In implementing this architecture, we made the following design decisions:

\mypara{Input volume generation.}
Given a 3D scene, we rotate it to align with gravity and room orientation based on Manhattan assumption.
The dimensions of the 3D space we consider are $4.8\unit{m}$ horizontally, $2.88\unit{m}$ vertically, and $4.8\unit{m}$ in depth.
We encode the 3D scene into a flipped TSDF with grid size $0.02\unit{m}$, truncation value $0.24\unit{m}$, resulting in a $240 \times 144 \times 240$ volume as the network input. 

\mypara{Capturing 3D context with big receptive field.}
Context can provide valuable information for understanding the scene, as demonstrated by much prior work in image segmentation \cite{YuKoltun2016}.
In the 3D domain, context is more useful due to a lack of high frequency signals compared to image textures.
For example, tabletops, beds, and floors are all geometrically similar to flat horizontal surfaces, so it is hard to distinguish them given only local geometry.
However, the relative positions of objects in the scene are a powerful discriminatory signal.
To learn this contextual information, we need to make sure our network has a big enough receptive field.
To this end, we extend the dilated convolution presented by Yu and Koltun~\cite{YuKoltun2016} to 3D.
Dilated convolution extends normal convolution by adding a step size when the convolution extracts values from the input before convolving with the kernel.
Thus we can exponentially expand the receptive field without a loss of resolution or coverage, while still using the same number of parameters.
\Cref{fig:RF_compare} compares the receptive field size of \OURS with 3D ConvNet architectures from prior work. 

\mypara{Multi-scale context aggregation.}
Different object categories have very different physical 3D sizes.
This implies that the network will need to capture information at different scales in order to recognize objects reliably.
For example, we need more local information to recognize smaller objects like TVs, while we need more global information to recognize bigger objects like beds.
In order to aggregate information at different scales we add a layer that concatenates the network responses with different receptive field.
We then feed this combined feature map into two $1\times1\times1$ convolution layers, which allows us to propagate information across responses from different scales.
%Because we use 3D voxel representation, the receptive fields directly reflect the physical size of 3D space 

\mypara{Data balancing.}
Due to the sparsity of 3D data, the ratio of empty vs. occupied voxels is around 9:1.
To deal with this imbalanced data distribution, we sample the training so that each mini-batch has a balanced set of empty and occupied examples.
For each training volume containing $N$ occupied voxels, we randomly sample $2N$ empty voxels from occluded regions for training. Voxels in free space, outside the field of view, or outside the room are ignored.

\mypara{Loss: voxel-wise softmax.}
The loss function of the network is the sum of voxel-wise softmax loss $L(p,y) = \sum\limits_{i,j,k}{w_{ijk}L_\mathrm{sm}(p_{ijk},y_{ijk})}$, where $L_\mathrm{sm}$ is softmax loss,  $y_{ijk}$ is the ground truth label, $p_{ijk}$ is the predicted probability of the voxel at coordinates $(i,j,k)$ over the $N+1$ classes, where $N$ is the number of object categories and empty voxels are labeled as class $0$.
The weight $w_{ijk}$ is equal to zero or one based on the sampling algorithm described above.

\mypara{Training protocol.}
We implement our network architecture in Caffe \cite{jia2014caffe}.
Pre-training \OURS on the SUNCG training set takes around a week on a Tesla K40 GPU, and fine-tuning on the NYU dataset takes 30 hours.
During training, each mini-batch contains one 3D view volume, requiring $11\unit{GB}$ of GPU memory.
To obtain more stable gradient estimates, we accumulate gradients over four iterations and update the weights once afterwards.

%We randomly initialize all layers by drawing weights from the Xavier algorithm \cite{glorot2010understanding}, and initialize biases to 0. We train with a fixed learning rate of $0.01$. We run SGD with a momentum of $0.99$, and weight decay of $0.0005$. 
%During training, each mini-batch has one volume, and we accumulate gradients over four iterations and update the weights once afterwards. Therefore our effective mini-batch size is four. 

% without weight updates, only updating the weights once afterwards.  %The learning rate is set to $0.01$. We run SGD with momentum of $0.99$, and parameter decay of $0.0005$. 

\section{Synthesizing training data}
\label{sec:dataset}

One of the main obstacles of training deep networks for scene-level dense 3D predictions is the lack of large annotated datasets with dense object semantic annotations at the voxel level.
Existing RGB-D datasets with surface reconstructions are subject to occlusions or partial observations, and cannot provide the volumetric occupancy and semantic labels for the entire space at the voxel level.
To obtain volumetric occupancy ground truth Firman \etal~\cite{FirmanCVPR2016} collect a tabletop dataset with reconstructed RGB-D video using KinectFusion~\cite{kinectfusion}. However, this data does not provide semantic labels, and only contains simple tabletop scenarios. 
In this paper, we present a new large-scale synthetic 3D scene dataset, from which we obtain a large amount of training data with synthetically rendered depth images and volumetric ground truth.

\subsection{SUNCG: a large-scale synthetic scene dataset}
%To create SUNCG we downloaded 3D scene models from the Planner5D website \cite{Planner5D}.
Our SUNCG dataset contains $45,622$ different scenes with realistic room and furniture layouts that are manually created though the Planner5D platform \cite{Planner5D}.
Planner5D is an online interior design interface that allows users to create multi-floor room layouts, add furniture from a object library, and arrange them in the rooms.
%The library consists of $2,644$ objects instances covering 84 object categories. %The diversity of the data mostly comes from different room layouts and object configurations. 
After removing duplicated and empty scenes, we ensured the quality of the data with a simple Mechanical Turk cleaning task.
During the task, we show a set of top view renderings of each floor and ask turkers to vote whether this is a valid apartment floor.
We collect three votes for each floor, and consider a floor valid when it has at least two positive votes.
In the end, we have $49,884$ valid floors, with contain $404,058$ rooms and $5,697,217$ object instances from $2644$ unique object meshes covering $84$ categories.
We manually labeled the all objects in the library to assign category labels.  \Cref{fig:data} shows example scenes from the resulting SUNCG dataset.
More information can be found in the appendix.

\begin{table*}[t]
\vspace{-3mm}
    \setlength{\tabcolsep}{4.3 pt}
    \centering\small
    \begin{tabular}{l|ccc|cccccccccccc}
    \toprule 
     & \multicolumn{3}{c|}{scene completion}  & \multicolumn{12}{c}{semantic scene completion } \tabularnewline
    %\cmidrule(lr){3-5}
    %\cmidrule(lr){6-17}
    \hline
    method (train) & prec. & recall & IoU & ceil.  & floor & wall & win. & chair & bed & sofa & table & tvs & furn. & objs. & avg. \tabularnewline
    \hline
    Lin \etal (NYU) \cite{lin2013holistic} & 58.5 &49.9 &36.4&0&11.7&13.3&{\bf 14.1}&9.4&29&24&6.0&7.0&16.2&1.1&12.0\tabularnewline
    Geiger and Wang (NYU) \cite{Geiger2015GCPR} & 65.7 &58&44.4&10.2&62.5&19.1&5.8&8.5&40.6&27.7&7.0&6.0&22.6&5.9&19.6\tabularnewline \hline
    \OURS (NYU) & 57.0&{\bf 94.5}&55.1&15.1&{\bf 94.7}&24.4&0&12.6&32.1&35&13&7.8&27.1&10.1&24.7\tabularnewline
    \OURS (SUNCG) & 55.6&91.9&53.2&5.8&81.8&19.6&5.4&12.9&34.4&26&13.6&6.1&9.4&7.4&20.2\tabularnewline
    \OURS (SUNCG+NYU) & {\bf 59.3}&92.9&{\bf 56.6}&{\bf 15.1} & 94.6 &{\bf 24.7} &10.8&{\bf 17.3} &{\bf 53.2} &{\bf 45.9}&{\bf 15.9}&{\bf 13.9}&{\bf 31.1} &{\bf 12.6} &{\bf 30.5}\tabularnewline 
    %\hline  \OURS  (SUNCG+ NYU)&render&74.2&97.0&72.5&38.7&92.9&47.7&9.1&31.1&56.5&57.8&32.4&11&43.2&26.3&40.6\tabularnewline
    %\hline
    \bottomrule
    \end{tabular}
    \caption{{\bf Semantic scene completion results on the NYU test set with kinect depth map.}}
    \label{table:full3dnyu}
\end{table*}

\subsection{Synthetic depth map generation}
To generate synthetic depth maps that mimic a typical image capturing process, we use a set of simple heuristics to pick camera viewpoints.
Given a 3D scene, we start with a uniform grid of locations spaced at $1\unit{m}$ intervals on the floor and not occupied by objects. We then choose camera poses based on the distribution of the NYU-Depth v2 dataset.\footnote{The camera height is sampled from a Gaussian distribution with  $\mu = 1.5\unit{m}$ and $\sigma = 0.1\unit{m}$. The camera tilt angle is sampled from a Gaussian distribution with $\mu = -10\degree$ and $\sigma = 5\degree$.}
Then, we render the depth map using the intrinsics and resolution of the Kinect.
After that we use a set of simple heuristics to exclude bad viewpoints.
Specifically, a rendered view is considered valid if it satisfies the following three criteria:
a) valid depth area (depth values in range of $1\unit{m}$ to $8\unit{m}$) larger than $70\%$ of image area,
b) there are more than two object categories apart from wall, ceiling, floor, and
c) object area apart from wall, ceiling, floor is larger than $30\%$ of image area.
To reduce data redundancy, we pick at most five images from each room.
In total we generate $130,269$ valid views for training our \OURS.

\subsection{Volumetric ground truth generation}
Since the 3D scenes in the SUNCG dataset consist of a limited number of object instances, we speed up the voxelization process by first voxelizing each individual object in the library and then transforming the labels based on each scene configuration and view point.
Specifically, we first voxelize each object to a ${128}\times{128}\times{128}$ voxel grid.
We set the voxel size $s$ so that the largest dimension of the object is a tight fit to the object bounding box.
Thus, $s$ varies between objects due to the difference in object dimensions.
We use the binvox~\cite{Binvox} voxelizer which accounts for both surface and interior voxels by using a space carving approach.

Given a camera view, we define a ${240}\times{144}\times{240}$ voxel grid in world coordinates, with scene voxel size equals to $2\unit{cm}$.
Then for each object in the scene, we transform the object voxel grid by translating, rotating and scaling by the object's transformation. % to obtain the set of voxel centroids in world coordinates.
We then iterate over each voxel in the scene voxel grid that is inside the transformed object bounding box, and calculate the distance to the nearest neighbor object voxel.
If the distance is smaller than the object voxel size $s$, this scene voxel will be labeled with this object category.
Similarly, we label all voxels in the scene that belong to walls, floors, and ceilings by treating them as planes with thickness equal to one scene voxel size.
All remaining voxels are marked as empty space, therefore providing a fully labeled voxel grid for the camera view.

\begin{figure*}[t]
    \vspace{-5mm}
    {\footnotesize
    ~~~~ RGB-D frame ~~~~~~~~ observed surface ~~~~~~~ground truth~~~~~~~~Zheng \etal \cite{ZhengZYIZ13Physics}~~~ Firman \etal\cite{FirmanCVPR2016}~~~~~Lin \etal\cite{lin2013holistic}~~ Geiger and Wang \cite{Geiger2015GCPR} ~~ \OURS\\
    }
    \vspace{-1.5mm}
    \includegraphics[width=\linewidth]{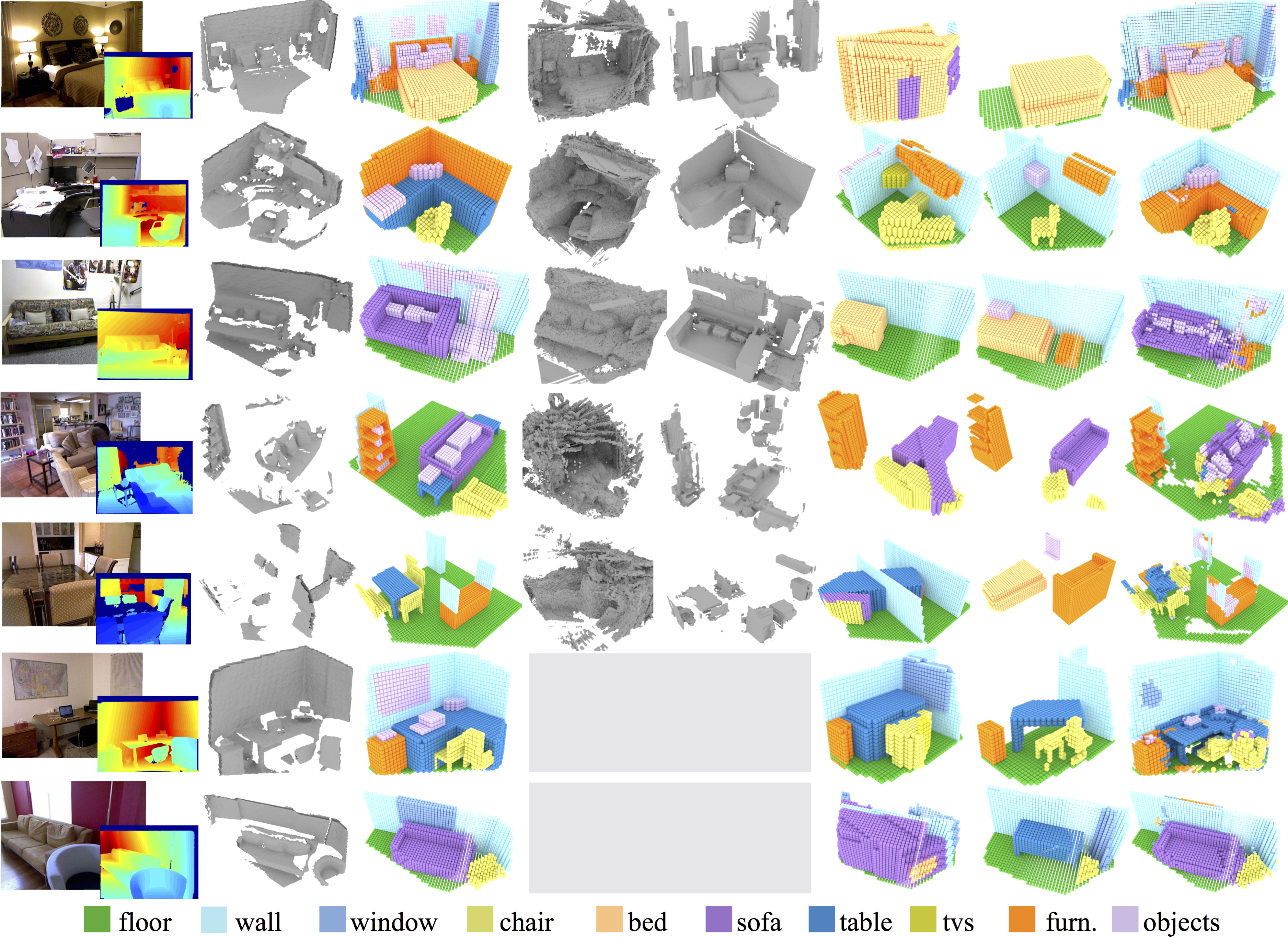}

    \caption{{\bf Qualitative results.} We compare with scene completion results from Zheng \etal~\cite{ZhengZYIZ13Physics} and Firman \etal~\cite{FirmanCVPR2016} (on a subset of the test set), and semantic scene completion results from Lin \etal~\cite{lin2013holistic} and Geiger and Wang~\cite{Geiger2015GCPR}. Zheng \etal \cite{FirmanCVPR2016} tested on rendered depth, \cite{lin2013holistic} and \cite{Geiger2015GCPR} tested on RGB-D frame from kinect, \cite{FirmanCVPR2016}  and \OURS tested on kinect depth. Overall, \OURS gives more accurate voxel predictions such as the lamps and pillows in the first row, and the sofa in the third row.}
    \label{fig:result}
\end{figure*}

%%%%%%%%%%%%%%%%%%%%%%%%%%%%%%%%%%%%%%%%%% Evaluation %%%%%%%%%%%%%%%%%%%%%%%%%%%%%%%%%%%%%%%%%%
\section{Evaluation}
In this section, we evaluate our proposed methods with a comparison to alternative approaches and 
an ablation study to better understand the proposed model.
We evaluate our algorithm on both real and synthetic datasets.

For the real data, we use the NYU dataset \cite{NYUdataset}, which contains 1449 depth maps captured from Kinect.
We obtain the ground truth by voxelizing the 3D mesh annotations from Guo \etal~\cite{guo2015predicting}, mapping object categories based on Handa \etal~\cite{SceneNet}.
The annotations consist of 33 object meshes in 7 categories, other categories approximated using 3D boxes or planes.
In some cases, the mesh annotation is not perfectly aligned with depth due to labeling error and the limited set of meshes. 
To deal with this misalignment, Firman at el.~\cite{FirmanCVPR2016} propose to use rendered depth map from the annotation for testing. However, by rendering the overly simplified meshes, geometric detail is lost especially in cases where objects are represented as a box.  Therefore, we test with both rendered depth maps and the originals.  

For synthetic data, we created a test set from SUNCG which has objects with detailed geometry, and for which the depth map and ground truth volumes are perfectly aligned. The SUNCG test set consists of 500 depth images rendered from 184 scenes that are not in the training set.

\mypara{Evaluation metric.}
As our evaluation metric, we use the voxel-level intersection over union (IoU) of predicted voxel labels compared to ground truth labels.
For the semantic scene completion task, we evaluate the IoU of each object classes on both the observed and occluded voxels.
For the scene completion task, we treat all non-empty object class as one category and evaluate IoU of the binary predictions on occluded voxels.
Following Firman \etal~\cite{FirmanCVPR2016}, we do not evaluate on voxels outside the view or the room.

\begin{table}
    \setlength{\tabcolsep}{6 pt}
    \centering\small
    \begin{tabular}{llccc}
    \toprule
    method & training & prec. & recall & IoU\tabularnewline
    \midrule
    Zheng \etal~\cite{ZhengZYIZ13Physics} & NYU & 60.1 & 46.7 & 34.6\tabularnewline
    Firman \etal~\cite{FirmanCVPR2016} & NYU & 66.5 & 69.7 & 50.8\tabularnewline
    \midrule
    \OURS completion & NYU &66.3  & 96.9 & 64.8\tabularnewline
    \OURS joint & NYU &  75.0& 	92.3& 	70.3 \tabularnewline
    \OURS joint & SUNCG+NYU  &  {\bf 75.0} & {\bf  96.0} & {\bf  73.0}\tabularnewline 
    \bottomrule
    \end{tabular}
    \caption{{\bf Scene completion on the rendered NYU test set as \cite{FirmanCVPR2016} }}
    \label{table:completion}
\end{table}

\subsection{Experimental results}
\Cref{table:completion,table:full3dnyu} summarize the quantitative results and \Cref{fig:result} shows qualitative comparisons. More qualitative results can be found in the appendix \Cref{moreresult,moreresult2}.

\mypara{Comparison to alternative approaches.}
In \Cref{table:full3dnyu} we compare on the semantic scene completion task with approaches from Lin \etal~\cite{lin2013holistic} and Geiger and Wang~\cite{Geiger2015GCPR}.
Both these algorithms take an RGB-D frame as input and produce object labels in the 3D scene.
Lin \etal use 3D bounding boxes and planes to approximate all objects.
Geiger and Wang retrieve and fit 3D mesh models to the observed depth map at test time.
The mesh model library used for retrieval is a superset of the models used for ground truth annotations.
Therefore, they can achieve perfect alignments by finding the exact mesh model in a small database.
In contrast, our algorithm is based on only depth and does not use additional mesh model at test time. 
Despite this data disparity, our network produces more accurate voxel-level predictions (30.5\% vs. 19.6\%).
An example of the difference is shown in the third row of \Cref{fig:result}: 
both Lin \etal and Geiger and Wang's approaches confuse the sofa as a bed while our network correctly recognizes the sofa.
Moreover, since our method does not require the model fitting step it is much faster at 7s compared to 127s per image \cite{Geiger2015GCPR}.
%Note that the over all low IoU number in this task also partially cased by misalignment between depth map and ground truth annotation as mention above. 

\mypara{Does recognizing objects help scene completion?}
Previous work has shown scene completion is possible without semantic understanding.
We examine to what extent the supervision of object semantics benefits the scene completion task.
To do this, we trained a model predicting the occupancy of each voxel by doing binary classification on each voxel (``empty'' vs. ``occupied'').
We compare the performance of models trained with occupancy and multi-class labeling (see \Cref{table:completion} [completion] vs. [joint]).
We also compare with Firmal \etal~\cite{FirmanCVPR2016} and Zheng \etal~\cite{ZhengZYIZ13Physics} which both predict binary voxel occupancy based on a single depth map without semantic understanding of the scene.
We use the re-implementation of Zheng \etal's approach from Firman \etal, which only provides the completion result.
We evaluate on the rendered NYU benchmark with the same test images used by Firman at al. (randomly picked 200 images from the full test set).
While Firman \etal produces good results for many cases, their approach fails when the scene becomes complex.
For instance, their algorithm fails to complete half of the bed in the first row of \Cref{fig:result}, and also fails to complete the chairs in the fifth row.
In contrast, \OURS is able to better complete the geometry by leveraging the semantics of the 3D context.
This result validates the idea that it is beneficial to understand object semantics in order to achieve better scene completion.

%our network is able to learn the complete 3D context while the observation of the data is always partial due to self-occlusion inter-object occlusion.
\begin{table*}[t]
\vspace{-6mm}
    \setlength{\tabcolsep}{3.6pt}
    \centering\small
    \begin{tabular}{lc|c|ccc|cccccccccccc}
    \toprule
     &  &  & \multicolumn{3}{c|}{scene completion} & \multicolumn{12}{c}{semantic scene completion}\tabularnewline
    \hline
    method  & encoding &  eval & prec. & recall & IoU & ceil. & floor & wall & win. & chair & bed & sofa & table & tvs & furn. & objs. & avg.\tabularnewline
    \hline
    no completion&flipped&observed&-&-&-&97.2&95.5&61.9&24.6&30.1&55.3&58.9&48.7&14.8&42.1&34.5&51.2\tabularnewline
    joint&flipped&surface&-&-&-&97.7&94.5&66.4&30.0&36.9&60.2&62.5&56.3&12.1&46.7&33.0&54.2\tabularnewline\hline
    no balancing&flipped&&73.1&95.8&70.8&{\bf 96.4}&85.3&52.1&25.8&16.5&47.1&45.7&28.1&15.3&37.1&19.8&42.7\tabularnewline
    Basic&flipped&observed&73.4&95.0&70.7&94.6&83.8&47.0&24.0&15.1&38.2&37.2&26.0&0.0&34.8&17.3&38.0\tabularnewline
    Basic+D&flipped&and&72.2&96.2&70.4&94.7&{\bf 85.9}&47.5&{\bf 29.2}&21.1&50.9&50.7&29.0&{\bf 21.3} &37.2&20.1&44.3\tabularnewline
    Basic+D+M&proj&occluded&72.0&92.3&67.9&91.6&80.9&45.1&14.6&10.2&39.4&29.8&19.8&0.0&27.4&14.3&33.9\tabularnewline
    Basic+D+M&tsdf&voxels&74.8&94.0&71.4&95.8&84.4&45.1&17.5&15.2&28.2&37.2&25.6&0.0&28.2&21.9&36.3\tabularnewline
    Basic+D+M&flipped&&{\bf 76.3}&{\bf 95.2}&{\bf73.5}&96.3&84.9&{\bf 56.8}&28.2&{\bf 21.3}&{\bf 56.0} &{\bf 52.7} &{\bf 33.7} &10.9&{\bf 44.3} &{\bf 25.4}&{\bf 46.4}\tabularnewline
    \bottomrule
    \end{tabular}
    \caption{{\bf Ablation study on SUNCG testset.} First two row shows the evaluation on surface segmentation with and without joint training.
    The following rows show the evaluation on semantic scene completion task. D: 3D dilated convolution. M: multi-scale aggregation.
    }\label{table:control}
\end{table*}

\mypara{Does scene completion help in recognizing objects?}
To answer this question, we trained a model with a loss only accounting for semantic labels evaluated on the visible surface and compared with the model trained jointly with labeling and completion (see \Cref{table:control} [no completion] vs. [joint]).
Even when only evaluating on the visible surface, the model trained with the added supervision of the scene completion task outperforms the model trained only on surface labeling ($54.2\%$ vs. $51.2\%$). This demonstrates that understanding complete object geometry and the 3D context is beneficial for recognizing objects.

\mypara{Does synthetic data help?}
To investigate the effect of using synthetic training data, we compared models trained only with NYU and models pre-trained on SUNCG and then fine-tuned on NYU (see \Cref{table:completion,table:full3dnyu} NYU vs. NYU+SUNCG).
We see a performance gain by using additional synthetic data especially for the semantic scene completion task having an $10.3\%$ improvement in IoU.
%Finetuning on NYU data helps on semantic labeling while not help much on shape completion, might due to the mis-alignment of the ground truth.

\begin{figure}[t]
    \includegraphics[width=\linewidth]{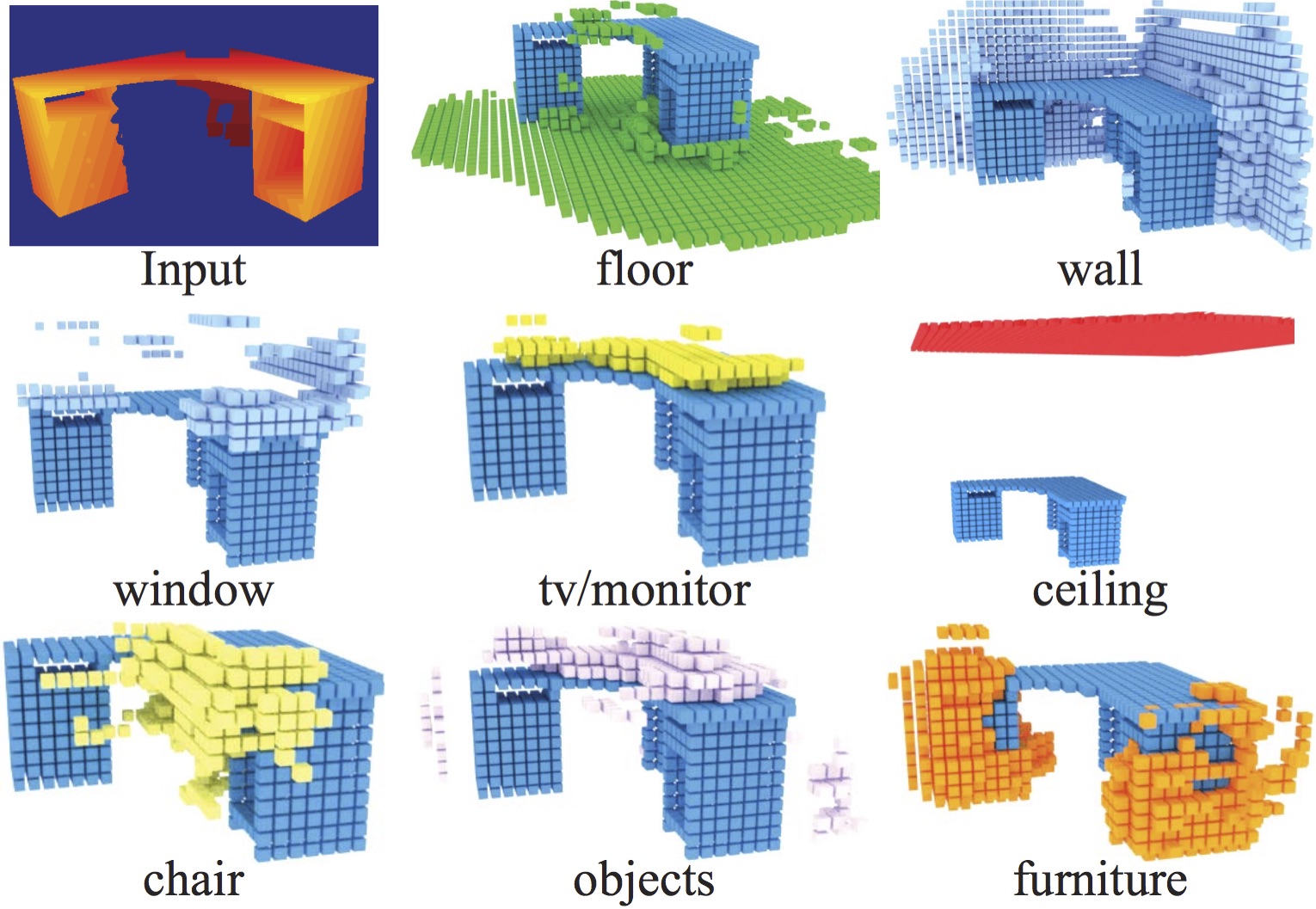}
    \caption{{\bf What 3D context does the network learn?} The first figure shows the input depth map (a desk) and the following figures show the predictions for other objects. Without observing any information for other objects the \OURS is able to hallucinate their locations based on the observed object and the learned 3D context.}\label{fig:context}
    \vspace{-3mm}
\end{figure}

\mypara{Does a bigger receptive field help?}
In \Cref{table:control}, the networks labeled [Basic] and [Basic+D] have the same number of parameter, while in [Basic+D] three convolution layers are replaced by dilated convolution, increasing the receptive field from $1.16\unit{m}$ to $2.26\unit{m}$.
Increasing the receptive field gives the network a opportunity to capture richer contextual information and significantly improve the network performance from $38.0\%$ to $44.3\%$. 
To visualize the contextual information learned by the network, we perform the following experiment: given a depth map of a single object we predict labels for all unobserved voxels.
\Cref{fig:context} shows the input depth and the predictions.
Even without observing depth information for other objects \OURS hallucinates plausible contextual object based on the observed object.
  
\mypara{Does multi-scale aggregation help?}
Comparing the network performance with and without the aggregation layer (see \Cref{table:control} [Basic+D] vs. [Basic+D+M]), we observe that the model with aggregation yields higher IoU for both the scene completion and semantic scene completion tasks by $3.1\%$ and $2.1\%$ respectively.

\mypara{Do different encodings matter?}
%In Section \ref{sec:encode} we discusses different ways of encoding the input geometry. 
The last three rows in \Cref{table:control} compare different volumetric encodings: projective TSDF [proj.], accurate TSDF [tsdf], and flipped TSDF [flipped].
We observe that removing the view dependency by using the accurate TSDF gives a $2.4\%$ improvement in IoU.
Making the gradients concentrated on the surface with the flipped TSDF leads to a $10.1\%$ improvement.

%At the same time, [fliiped proj.] achieve a very close performance as [fliiped], which demonstrate that have the strong gradient on surface is more important than remove the view dependency.

\mypara{Is data balancing necessary?}
To balance the empty and occupied voxel examples, we proposed to sample the empty voxels during training.
In \Cref{table:control}, [no balancing] shows the performance when we remove the sampling process during training, where we see a  drop in IoU  from $46.4\%$ to $42.7\%$.
%This leads to a drop from $46.4\%$ to $42.7\%$.

\mypara{Limitations.}
Firstly, we do not use any color information, so objects  missing depth  such as ``windows'' are hard to handle.
This also leads to confusion between objects with similar geometry or functionality.
For example, in the second row of \Cref{fig:result} the network predicts the desk as the broader furniture category.
Secondly, due to the GPU memory constraints, our network output resolution is lower than that of input volume.
This results in less detailed geometry and missing small objects, such as the missed objects on the desk of the second row in \Cref{fig:result}.

\section{Conclusion}
In this paper, we introduced \OURS, a 3D ConvNet for the semantic scene completion task of jointly predicting volumetric occupancy and semantic labels for full 3D scenes. We trained this network on a new large-scale synthetic 3D scene dataset.
Experiment results demonstrate that our joint model outperforms methods addressing either component task in isolation, and that by leveraging the 3D contextual information and the synthetic training data, we significantly outperform alternative approaches on the semantic scene completion task.

\begin{figure*}[t]
\vspace{-5mm}
    \includegraphics[width=\linewidth]{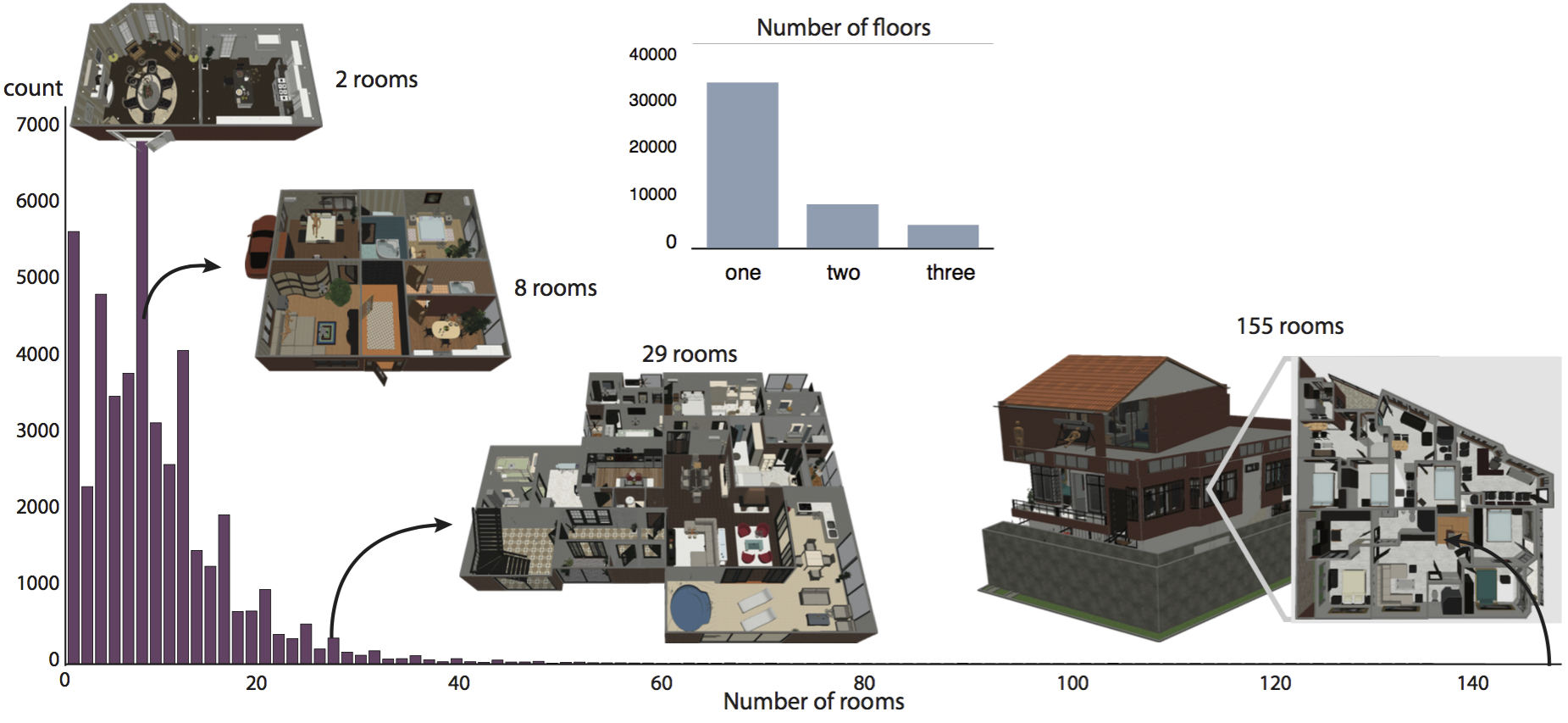}
    \caption{\label{room_dis} {\bf Scene structure statistics.} Distribution of number of rooms and number of floors in our SUNCG dataset.
    Our dataset contains large variety of 3D indoor scenes such as small studios, multi-room apartments, and multi-floor houses.}
\end{figure*} 

\mypara{Acknowledgment}
This work is supported by Intel, Adobe, and NSF (IIS-1251217 and VEC 1539014/ 1539099).  It makes use of data from Planner5D and hardware donated by NVIDIA and Intel. Shuran Song is supported by a Facebook Fellowship.   Helpful advice was provided by Michael Firman, Yinda Zhang, Matthias Niessner and Angela Dai.

%%%%%%%%%%%%%%%%%%%%%%%%%%%%%%%%%%%%%%%%%% Appendix %%%%%%%%%%%%%%%%%%%%%%%%%%%%%%%%%%%%%%%%%%

\appendix

\section{SUNCG Dataset Statistics}
In this section, we present several statistics related to our SUNCG dataset. We start by providing the basic statistics of scene structure and physical size for 3D scenes in our dataset, and then move on to talk about higher-level statistics regarding object categories, room types, and object-room relationships.

\begin{figure*}[t]
    \includegraphics[width=\linewidth]{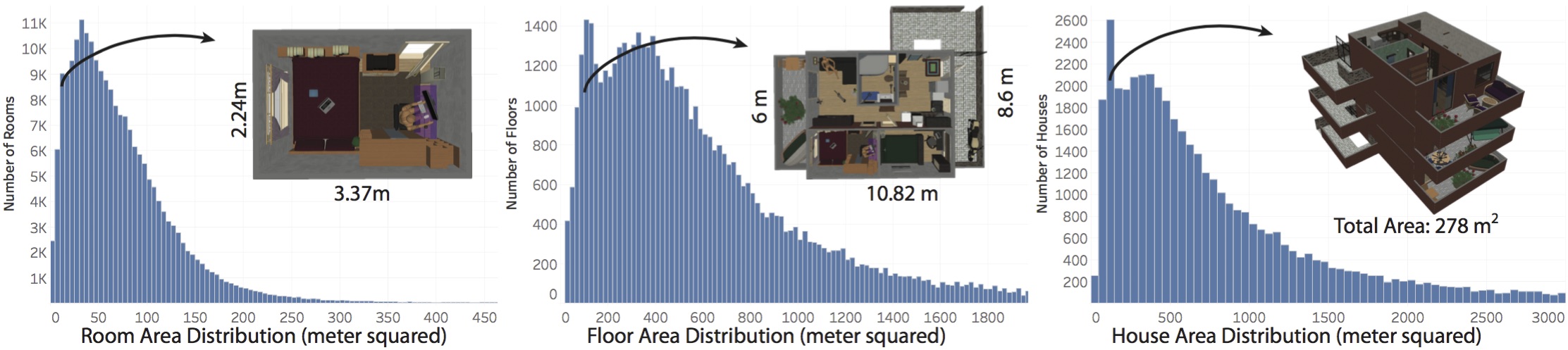}
    \caption{\label{physical_size} {\bf Distribution of physical sizes} (in meters\textsuperscript{2}) per room, floor, and house of the SUNCG dataset.}
\end{figure*}

\paragraph{Scene Structure Statistics}
Figure \ref{room_dis} illustrates the distribution of number of rooms and number of floors per scene in the SUNCG dataset.
The 3D scenes in our dataset are range from single room studio to multi-floor houses. 
The average and median number of rooms per-house are 8.9 and 7 respectively. The average and median number of floors per-house are 1.3 and 1 respectively.

\paragraph{Physical Size Statistics}
All object meshes and 3D scenes in the SUNCG dataset are measured in real-world spatial dimensions (units are in meters). 
Figure \ref{physical_size} shows statistics related to physical size over three levels: rooms, floors and houses.

\begin{figure*}[t]

    \includegraphics[width=\linewidth]{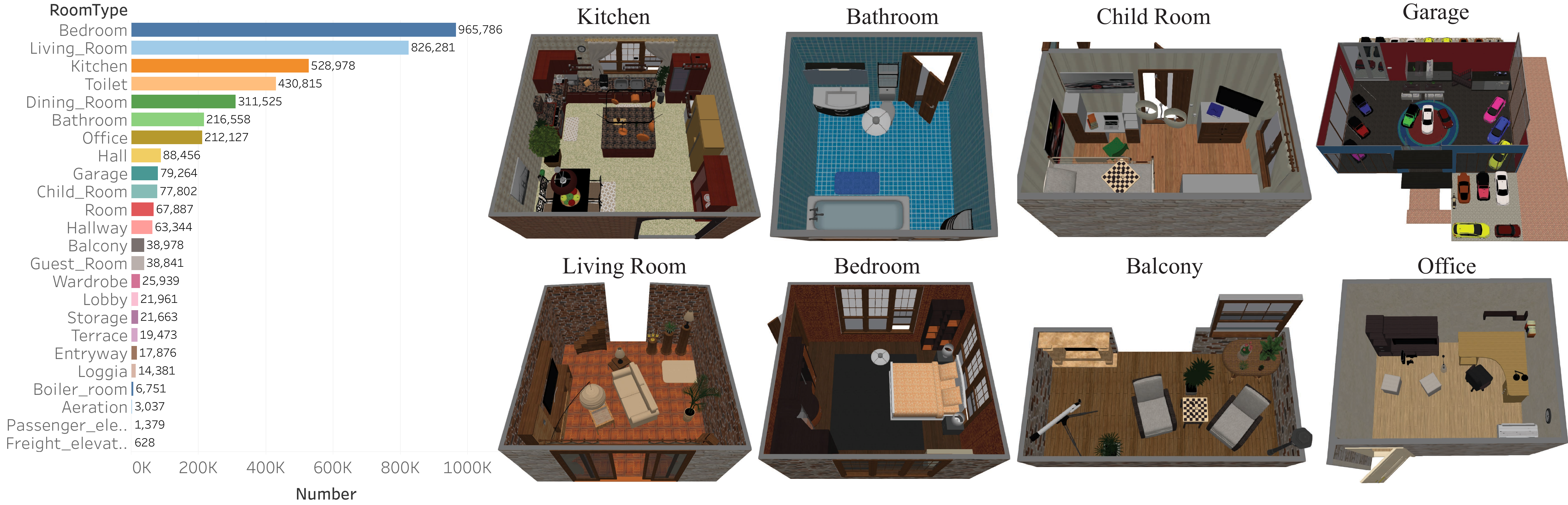}
    \caption{\label{roomtype} Distribution of different room types in the SUNCG dataset (left), and examples of rooms per room type (right).}
 \end{figure*} 
%\vspace{6mm}
\begin{figure*}[t]
    \includegraphics[width=\linewidth]{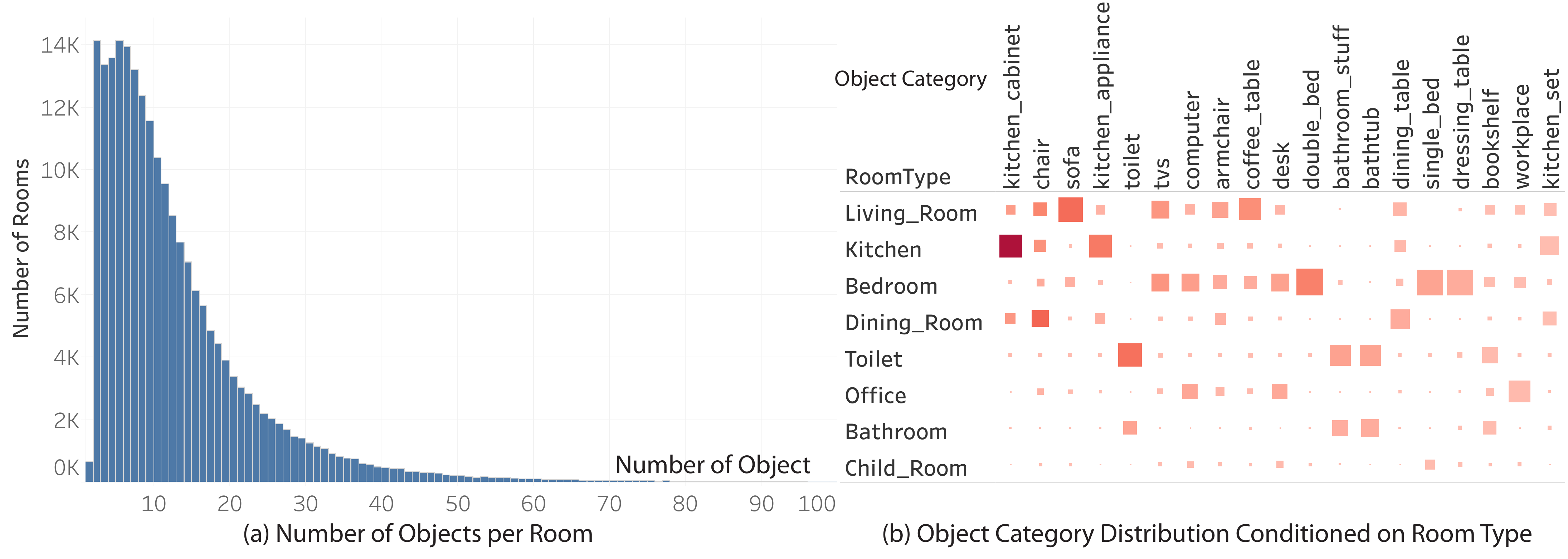}
    \caption{\label{obj_room} {\bf Object-Room Relationship.} 
    On the left we show the distribution of number of objects in each room. On average there are more than 14 objects in each room. 
    On the right we show the object category distribution conditioned on different room type. Size of the square shows the frequency of given object  category appears in the certain room type.  The frequency is normalized for each object category.
    As expected,  object occurrences are tightly correlated with the room type. For example, kitchen counters has a very high chance to appear in kitchen, tvs are more likely to appear in living room or bedroom, while chairs appear frequently in many room types.
    }
\end{figure*}

\paragraph{Room Type Statistics}
Figure \ref{roomtype} shows the room type distribution and several example rooms per type from our dataset.
In total, we have 24 room types that are labeled by the user during creation. These labels include:
living room, kitchen, bedroom, child room, dining room, bathroom, toilet, hall, hallway, office, guest room, wardrobe, room, lobby, storage, boiler room, balcony, loggia, terrace, entryway, passenger elevator, freight elevator, aeration, and garage. 
The four most common room types in our dataset are bedroom, living room, kitchen and toilet, which agrees with the distribution in real-world living spaces.

\begin{figure*}[t]
    \includegraphics[width=\linewidth]{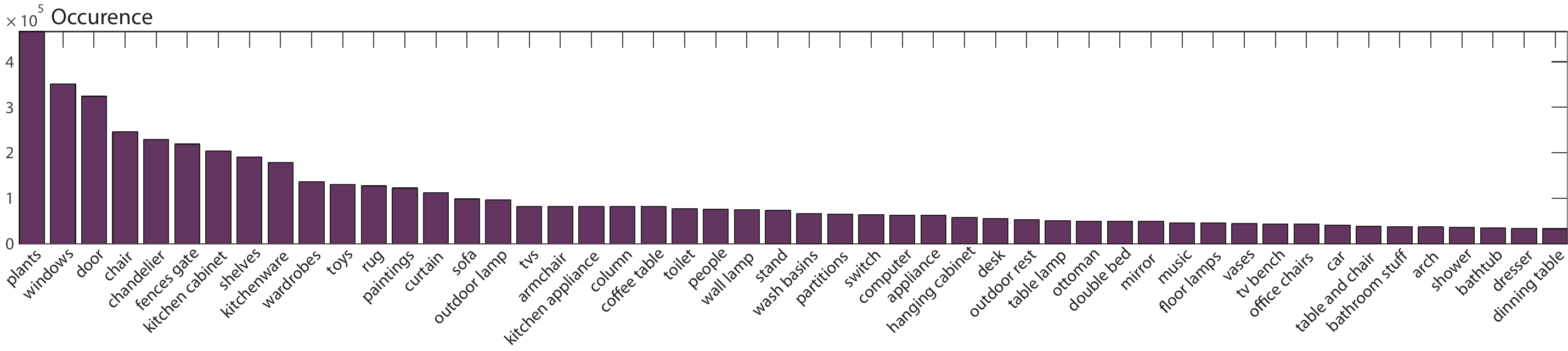}
    \caption{\label{sup_objectscate_dis} {\bf Distribution of object categories in the SUNCG dataset.} We have 84 object categories in total. Here we show the top 50 object categories with highest number of occurrences in our dataset.}
    % \end{figure*} 
    \vspace{3mm}
    % \begin{figure*}[t]
    \includegraphics[width=\linewidth]{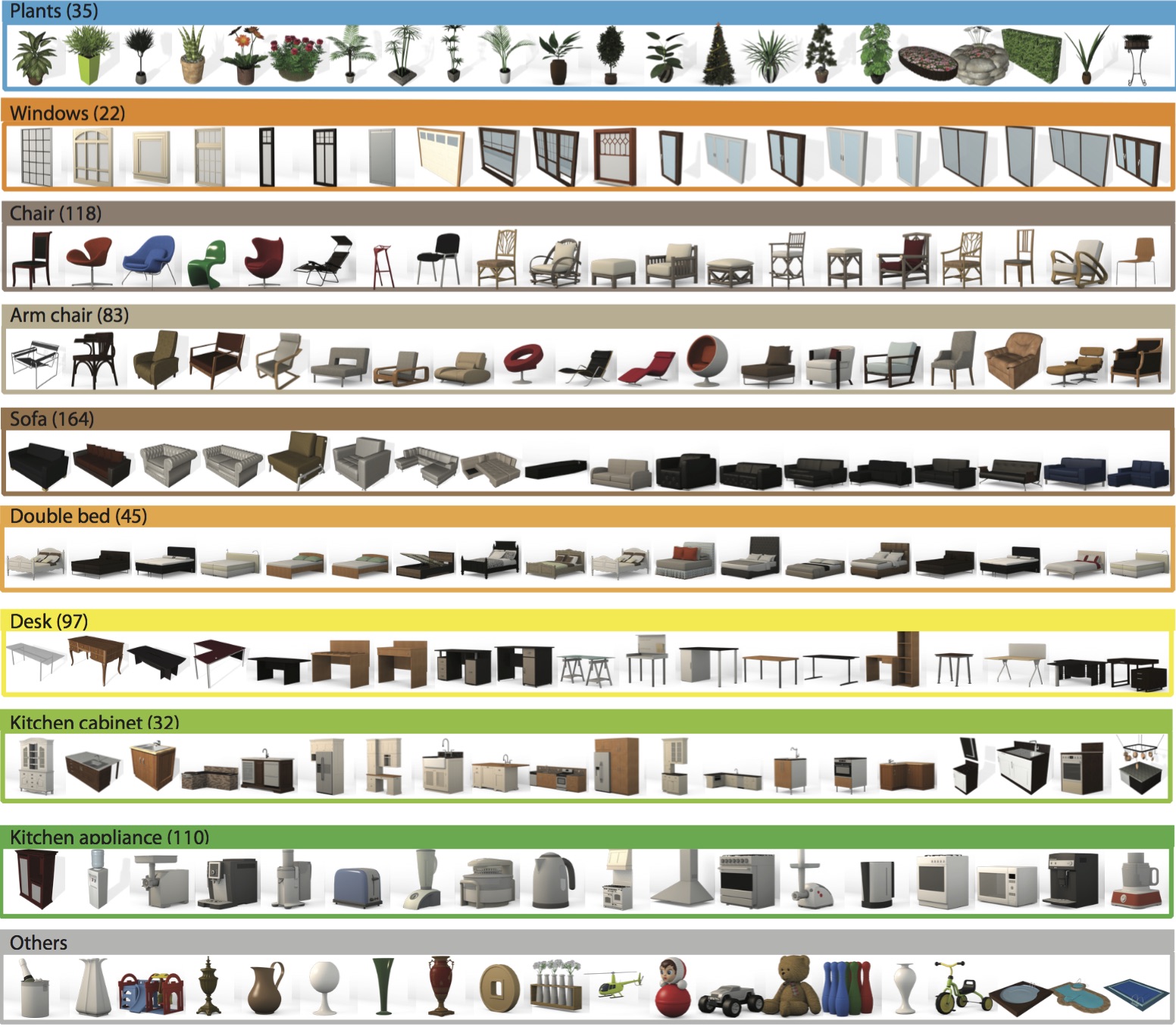}
    \caption{\label{sup_objects} {\bf Object meshes of selected object categories.} The total number of unique object meshes per category in the object library is listed in parentheses. During the creation of 3D scenes, users have the flexibility to reshape, resize, and reapply textures to objects to better fit the room arrangement and style, which further improves the diversity of our dataset. }
\end{figure*} 

\paragraph{Object Category Statistics}
Figure \ref{sup_objectscate_dis} shows overall object category occurrence in the SUNCG dataset.
Figure \ref{sup_objects} shows examples of object models from the object library, which contains a diverse set of common furniture and objects for common living spaces. Furthermore, during the creation of the 3D scenes, users have the flexibility to reshape, resize, and re-apply texture to objects to better fit the room style, which further improves the dataset diversity.

\paragraph{Object-Room Relationships}
With complete object and room type annotations, we can further study the object-room relationships in our dataset.
Figure \ref{obj_room} (a) shows the distribution of number of objects per room.
Figure \ref{obj_room} (b) shows the distribution of object categories conditioned on different room types. 
On average there are more than 14 objects in each room. 
The occurrence and arrangements of these objects in rooms provide rich contextual information that we can learn from.

\begin{figure*}[t]
    \includegraphics[width=\linewidth]{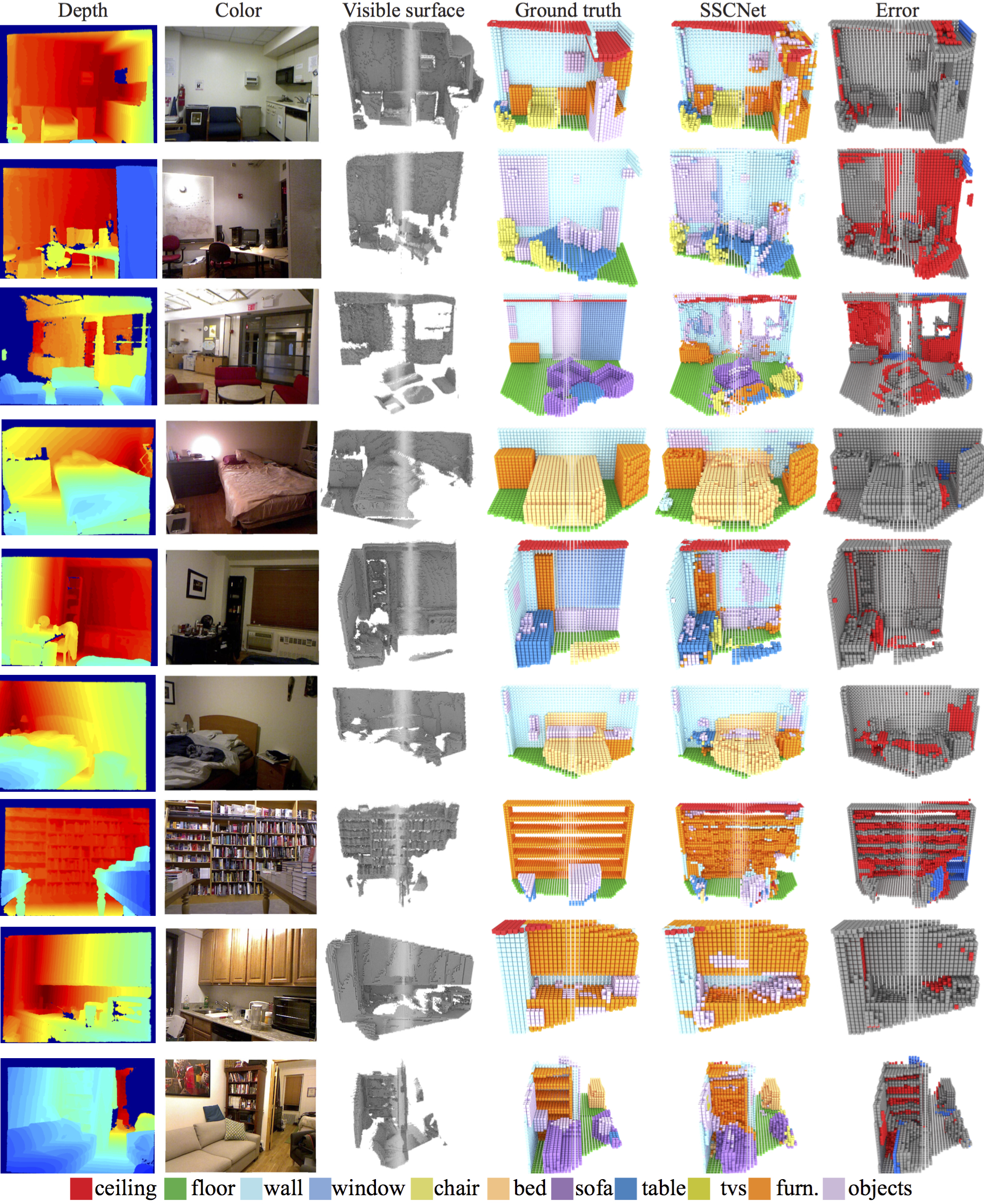}
    \caption{\label{moreresult} {\bf Results and error visualization.} 
    The first three columns show the input depth map, corresponding color image and visible surface. 
    The fourth and fifth columns show the ground truth and prediction results for the semantic scene completion task. 
    The sixth column visualizes the error of completion result in the evaluation region (not include voxels in observed free space, or outside field of view). Gray voxels indicates true positive, red voxels indicates false positive and blue indicates false negative for the binary scene completion task.}
\end{figure*} 

\begin{figure*}[t]
    \includegraphics[width=\linewidth]{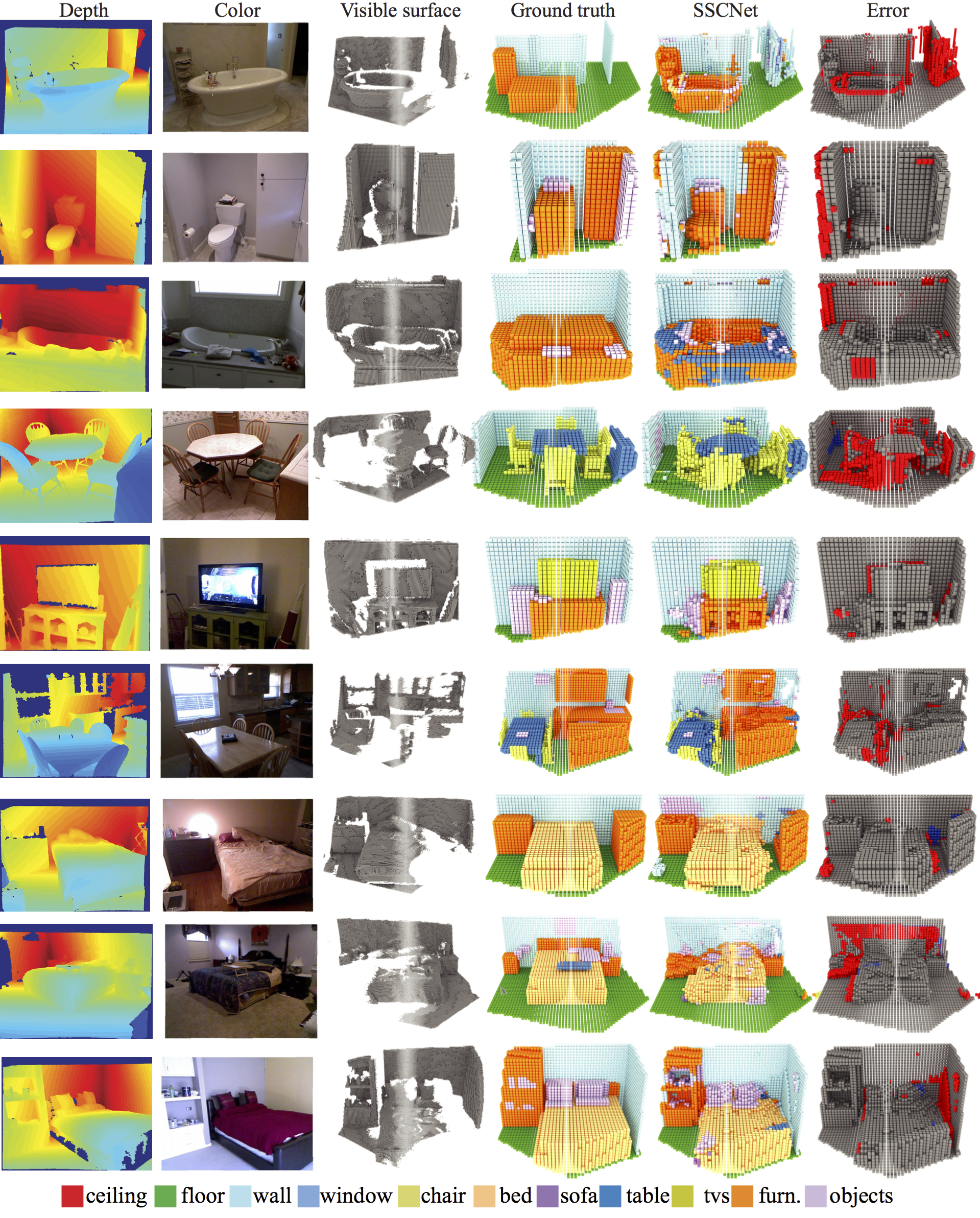}
    \caption{\label{moreresult2} {\bf Results and error visualization.} see Figure \ref{moreresult}. }
    \label{fig:dataset_supp}
\end{figure*}

% \section{Implementation Details}
% We implement our network architecture in Caffe \cite{jia2014caffe}.
% We randomly initialize all layers by drawing weights from the Xavier algorithm \cite{glorot2010understanding}, and initialize biases to 0. We train with a fixed learning rate of $0.01$. We run SGD with a momentum of $0.99$, and weight decay of $0.0005$. 
% During training, each mini-batch has one volume, and we accumulate gradients over four iterations and update the weights once afterwards. Therefore our effective mini-batch size is four. 
% Figure \ref{moreresult}  and \ref{moreresult2} show more results and error visualizations.

%\section{Additional Results}
% Table \ref{table:surf} shows 2D surface segmentation results and comparisons with Handa \etal \cite{SceneNet}. Handa \etal~ synthetically rendered 10K depth maps from a small synthetic scene dataset and use them to pretrained their 2D segmentation network based on the HHA encoding of the depth map.
% For our model, the 2D surface segmentation is obtained by back-projecting the predicted 3D voxel labels on visible surface onto 2D image plane.
% We evaluate 2D pixel-level intersection over union with the same set of object categories as Handa \etal \cite{SceneNet}.
% Our method achieves very close performance in comparison to Handa \etal \cite{SceneNet}. However, unlike Handa \etal \cite{SceneNet}, we directly train our model on the synthetic dataset without using pre-trained weights from ImageNet.
% Figure \ref{moreresult}  and \ref{moreresult2} show more results and error visualizations on NYU testset.

{\small
\bibliographystyle{ieee}
\bibliography{RGBDdetect,pvg,ivs-eccv16}
}

\end{document}